\pdfoutput=1

\documentclass[11pt]{article}

\usepackage[]{acl}

\usepackage{times}
\usepackage{latexsym}

\usepackage[T1]{fontenc}

\usepackage[utf8]{inputenc}

\usepackage{microtype}

\usepackage{subcaption}
\usepackage{caption}
\usepackage{multirow}

\usepackage{amssymb}
\usepackage{pifont}

\usepackage{adjustbox}
\usepackage{array, makecell}

\usepackage{amsmath}
\usepackage{graphicx}
\usepackage{subcaption}
\usepackage{caption}

\definecolor{text}{rgb}{0,1,0}
\definecolor{title}{rgb}{1,0,0}
\definecolor{list}{rgb}{1,1,0}
\definecolor{table}{rgb}{0.37,0.15,0.07}
\definecolor{figure}{rgb}{0.25,0.41,0.88}

\definecolor{question}{rgb}{0,1,0}
\definecolor{answer}{rgb}{1,0,0}
\definecolor{header}{rgb}{1,1,0}
\definecolor{other}{rgb}{0.25,0.41,0.88}
\definecolor{abstract}{rgb}{0.75,0.22,1}
\definecolor{author}{rgb}{0.6,1,1}
\definecolor{caption}{rgb}{0,0,1}
\definecolor{date}{rgb}{0.5,0.5,0.5}
\definecolor{equation}{rgb}{0.4,0.22,1}
\definecolor{figure_doc}{rgb}{0.25,0.41,0.88}
\definecolor{footer}{rgb}{1,0.6,0}
\definecolor{list_doc}{rgb}{1,1,0}
\definecolor{paragraph}{rgb}{0,1,0}
\definecolor{reference}{rgb}{0,1,1}
\definecolor{section}{rgb}{1,0,0}
\definecolor{table_doc}{rgb}{0.37,0.15,0.07}
\definecolor{title_doc}{rgb}{1,0.5,0.5}

\title{Doc-GCN: Heterogeneous Graph Convolutional Networks for Document Layout Analysis}
\author{
Siwen Luo\textsuperscript{1\thanks 
{
 ~~Co-First Authors
}}, 
Yihao Ding \textsuperscript{1*},
Siqu Long\textsuperscript{1}, 
Josiah Poon\textsuperscript{1},
Soyeon Caren Han\textsuperscript{1, 2\thanks{
~~Corresponding Author (caren.han@sydney.edu.au)
}}
\\
\textsuperscript{1}The University of Sydney, \textsuperscript{2} The University of Western Australia\\
\tt\small 
\{siwen.luo, yihao.ding, josiah.poon, caren.han\}@sydney.edu.au,\\ 
\tt\small 
caren.han@uwa.edu.au, slon6753@uni.sydney.edu.au\\
}

\begin{document}
\maketitle
\begin{abstract}
Recognizing the layout of unstructured digital documents is crucial when parsing the documents into the structured, machine-readable format for downstream applications. Recent studies in Document Layout Analysis usually rely on visual cues to understand documents while ignoring other information, such as contextual information or the relationships between document layout components, which are vital to boost better layout analysis performance. Our Doc-GCN presents an effective way to harmonize and integrate heterogeneous aspects for Document Layout Analysis. We construct different graphs to capture the four main features aspects of document layout components, including syntactic, semantic, density, and appearance features. Then, we apply graph convolutional networks to enhance each aspect of features and apply the node-level pooling for integration. Finally, we concatenate features of all aspects and feed them into the 2-layer MLPs for document layout component classification. Our Doc-GCN achieves state-of-the-art results on three widely used DLA datasets: PubLayNet, FUNSD, and DocBank. The code will be released at \url{https://github.com/adlnlp/doc_gcn}
\end{abstract}

\section{Introduction}
Digital documents (incl. Scanned Document Images and PDF files) are popular and convenient for storing written textual information, so almost 2.5 trillion documents worldwide are available in the digital format~\cite{zhong2019publaynet}. However, it is challenging to automatically recognize the layout and components of these unstructured digital documents and extract meaningful information using this format. For example, the financial office would require the scanned document image after their client signed. It is then crucial to recognize and extract the form component, such as the form title, person name, and the date the document is signed. This task is widely called Document Layout Analysis (DLA). The DLA task aims at understanding the documents from either 1) the physical analysis by detecting the document structure and the boundaries of each layout region or 2) the logical analysis by categorizing the detected layout components (segments) into the predefined document element classes, such as Title, Date, Author, and Figure~\citep{binmakhashen2019document}. In this research, we focus on the logical DLA task to classify the different layout components of PDF documents by understanding the relationships between components. Traditional deep learning-based DLA approaches mainly focus on processing visual features of layout components~\citep{soto2019visual,augusto2017fast,li2020cross} using CNN-based models. Some recent studies started to use texts to solve the problem with the support of semantic information for each layout component~\citep{li2020docbank,xu2020layoutlm}. However, applying visual and textual features is not enough to analyze the characteristics and relations of document layout components in order to classify them. In this paper, we try to fill this gap by defining and proposing: 1) Layout Components Characteristic Representation and 2) Relation Representation between components.

The first question would be \textit{`What would be the best aspects to represent the characteristics of different document layout components?'}. The text density/sparsity in each document component is a valuable feature. For example, a \textit{paragraph} is more dense and usually contains more texts than a \textit{table}. Moreover, syntactic information can be a key characteristic. It is obvious that a \textit{title} mainly consists of noun phrases only, whereas a paragraph contains more sentences with the noun and verb phrases. Hence, in this research, apart from the common visual features and semantic text features used by previous works, we propose the four major aspects, including text density, components appearance, syntactic and semantic information of textual contents, of each component in order to let the model conduct more comprehensive learning of the Documents properties. Another question would be \textit{`How to represent the relation among document layout components?'} Most DLA studies did not apply multiple aspects of features. Even if applied~\citep{soto2019visual,xu2020layoutlm}, those features are integrated via simple concatenations and do not consider the influences of relationships between layout components on the classification performances. The characteristics of each component/segment are not enough to analyze the whole document layout and its corresponding components. Assume a `text' component is detected based on its characteristics, but it is more accurate to classify it as the `figure caption' if it is right above or below a figure. Thus, capturing and encoding the relationships between layout components is crucial for better layout analysis.

In this paper, we propose a novel Heterogeneous Graph Convolutional Network (GCN)-based DLA model, Doc-GCN, on a document page level, taking the document layout component (segment) as nodes in the graph and encoding the relative positional and structural relations between layout components. We first construct six different graphs, each encoding one aspect of features among the layout components' syntactic, semantic, text density, and visual features. The syntactic and density aspects have two graph variants based on the different node embedding initialization methods. We use the GCN to update the node embedding by learning from its intimate neighbors, and the node-level pooling is then applied to integrate the graph variants. We concatenate the updated features of each aspect, getting the final layout component representation for the final classification of layout component types. 

In summary, the contributions of our work are as follows:
\begin{itemize}
    \item To the best of our knowledge, this is the first attempt to apply heterogeneous aspects of Document Layout Analysis. 
    \item Doc-GCN is the first to propose using multi-aspect Graph Convolutional Networks for harmonizing the characteristics and relationships among document layout components (segments). 
    \item Doc-GCN achieved the state-of-the-art performance on three widely used DLA datasets, PubLayNet, FUNSD, and DocBank.
\end{itemize}

\section{Related Work}
\subsection{Document Layout Analysis}
In the 1990s, rule-based methods \cite{klink2001rule,fisher1990rule,niyogi1986rule} were widely used for the DLA tasks until the rise of deep learning. \citet{zhong2019publaynet} used Faster RCNN~\cite{ren2015faster} and Mask RCNN~\cite{he2017mask} as the basic deep learning models for DLA task. Recent works have extensions. ~\citet{soto2019visual} added the size of the proposed Region of Interest (ROI) and normalized page number as the additional contextual information to the pooled feature vectors for both classification and regression network of Faster RCNN. ~\citet{augusto2017fast} proposed the 1-D CNN with the parallel operation of horizontal and vertical projection. ~\citet{li2020cross} proposed a model based on the Feature Pyramid Networks (FPN) object detector to solve the cross-domain document layout classifications.

Some DLA works integrated not only visual features but also textual features. ~\citet{xu2020layoutlm} proposed a pretrained model that integrated each token's positional and text embeddings with the corresponding image embeddings. \citet{xu-etal-2021-layoutlmv2} then applied additional pretraining tasks to enhance the multi-modality interactions further and used a spatial-aware attention mechanism to capture the relative positional relationship between different layout components. \citet{gu2021unidoc} proposed the pretraining framework with a cross-attention transformer to boost the more substantial alignment between visual and textual features for each document element region. \citet{li2020docbank} proposed a new dataset on the token-level where each token is annotated into a layout element class and experimented with this dataset with pretrained language models: BERT~\cite{devlin2019bert} and RoBERTa~\cite{liu2019roberta} by inputting the sequence of token embeddings with the corresponding bounding boxes. \citet{10.1007/978-3-030-86549-8_8} used a relation module upon the multimodal representations integrated from vision and semantic features to detect the relations between different components. 

However, understanding the vision and semantic aspects of individual objects/regions is not enough to analyze the document layout and components. It is critical to consider the influences of relationships between document layout components. This is very similar to the trend that we can see in the most visual-language tasks~\cite{ijcai2022p773}. For example, the table caption should be around the table, and each component should be semantically related. Besides, it is also important to provide the better interpretation of the DLA prediction semantically, which has been considered as lot of NLP tasks~\cite{luo2021local}.

\subsection{Graph Convolutional Networks}
Graph Convolutional Networks (GCN) \cite{kipf2016semi} is a type of Graph Neural Network which applies convolution over graph structures, and it has been applied to many Natural Language Processing (NLP) tasks. For example, TextGCN~\cite{yao2019graph} and MEGCN~\cite{wang2022me} focus on the text classification by representing words and documents as graph nodes, and TensorGCN~\cite{liu2020tensor} captures their relations in different aspects, including semantic, syntactic, and sequential aspects. D-GCN~\cite{chen2020joint} performs sentiment analysis jointly with aspect extraction for graph-based modeling. Hier-GCN \cite{cai2020aspect} proposes a hierarchical GCN to model the inner- and inter-relations among multiple aspect categories and sentiments. InducT-GCN~\cite{wang2022induct} enables the inductive GCN learning model, which improves the performance and reduces the time complexicity. 

Such graph-based approach is also widely applied to multimodal tasks, especially for visual question answering \cite{huang2020aligned,luo2020rexup}, text-to-image generation \cite{johnson2018image,han2020victr}, and text-image matching \cite{liu2020graph,long2022gradual}. It receives lots of attention in converting multiple modalities and aspects into structured graphs and enhancing joint learning. Some document-based analysis works, such as document dating \cite{vashishth2018dating}, apply the GCN-based document dating approach by jointly exploiting the document's syntactic and temporal graph structures. \cite{wang2022post} also uses two-stage GCN classifiers for line splitting and clustering for paragraph recognition in documents. In this work, we apply GCN to the DLA task by joint-learning different aspects of document layouts and capturing the relationship between layout components.

\begin{figure*}[t]
\centering
\includegraphics[width=\linewidth]{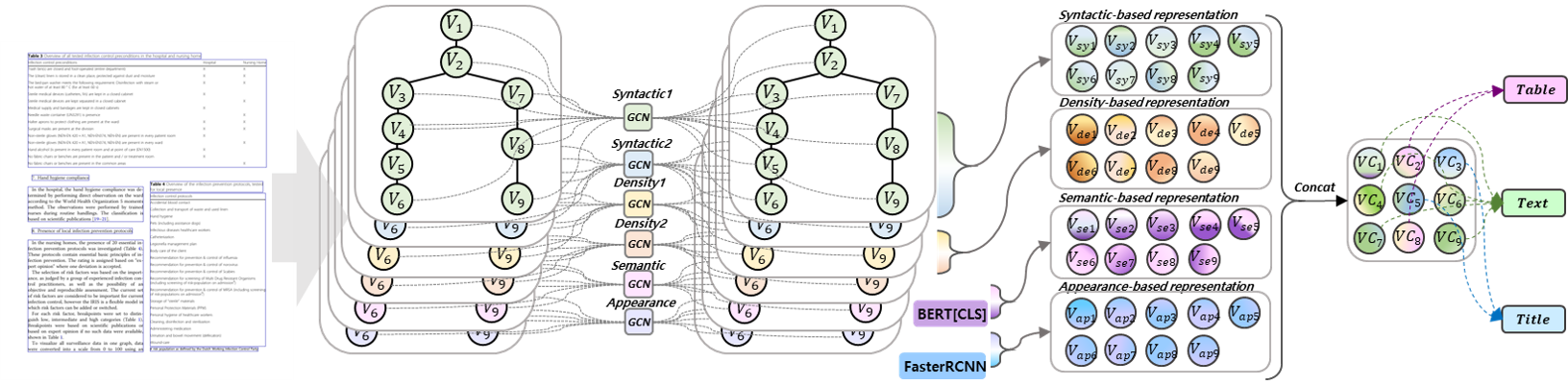}
\caption{The overall architecture of the proposed Doc-GCN for document layout component classification. }
\label{fig:rep_cons}
\end{figure*}

\section{Doc-GCN}\label{sec:docgcn}
We propose a graph-based network Doc-GCN to encode and integrate the different aspects of document layout components. For each PDF page with $N$ document layout components, we construct six different graphs: $\mathcal{G}_{den1}$, $\mathcal{G}_{den2}$, $\mathcal{G}_{appr}$, $\mathcal{G}_{syn1}$, $\mathcal{G}_{syn2}$ and $\mathcal{G}_{semc}$, capturing the features of all the document layout components in the page from four aspects: density, appearance, syntactic and semantic. Each graph $\mathcal{G}_{i} = \left ( \mathcal{V}_i, \mathcal{E}_i, \mathcal{A}_i \right )$ consists of a set of nodes $\mathcal{V}_{i}\left (\left | \mathcal{V}_{i} \right |=N  \right )$, a set of edges $\mathcal{E}_{i}$, and an adjacency matrix $\mathcal{A}_{i} \in \mathbb{R}^{N \times N}$. We regard each layout component of the page as a node $\upsilon_{n} \in \mathcal{V}_{i}$ in a graph. We segment each layout component by its bounding box. Different node embedding initialization and edge connection methods are applied for different graphs to match the characteristics of different feature aspects and capture different node relationships. The Graph Convolution Network (GCN) is then trained to update the node embeddings by learning from the neighbors. 

\subsection{Preliminaries}\label{sec:GCN}
GCN~\cite{kipf2016semi} is a convolutional neural network that operates directly on a graph to update the node embeddings by learning from the neighbors of each node. Given an initial node embedding matrix $\mathcal{H}^{0} \in \mathcal{R}^{N \times d_{0}}$ consisting of $N$ node features of size $d_{0}$, GCN will conduct the propagation through layers based on the rule in Equation~\ref{gcn}.
\begin{equation}\label{gcn}
    \mathcal{H}^{l+1} = f\left ( \mathcal{H}^{l}, \mathcal{A}\right ) = \sigma \left (\hat{\mathcal{A}}\mathcal{H}^{l}W^{l} \right )
\end{equation}
The node embedding matrix will be updated from $\mathcal{H}^{l} \in \mathbb{R}^{N \times d_{l}}$ to $\mathcal{H}^{l+1} \in \mathbb{R}^{N \times d_{l+1}}$ after every GCN layer, where $l = 1,2,...,L$ indicates the layer number. $\hat{\mathcal{A}} = \widetilde{\mathcal{D}}^{-\frac{1}{2}}\widetilde{\mathcal{A}}\widetilde{\mathcal{D}}^{\frac{1}{2}}$ represents the normalized symmetric adjacency matrix where $\widetilde{\mathcal{A}}=\mathcal{A}+\mathcal{I}$ and $\mathcal{I}$ is the identity matrix for self-connection inclusion. $\widetilde{\mathcal{D}}$ is the diagonal node degree matrix with $\widetilde{\mathcal{D}}(i,i)=\Sigma_j\widetilde{\mathcal{A}}(i,j)$ and $W^{l} \in \mathcal{R}^{d_{l} \times d_{l+1}}$ is the trainable weight matrix associated with the $l$-th layer. $\sigma$ denotes the activation function that can be different for different GCN layers. 

\subsection{Graphs Construction} \label{sec: graph_construction}

\subsubsection{Density-aspect Graph\footnote{\label{den_app}Details of density and appearance graph construction procedure can be found in Appendix A.1}}\label{subsec:density}
Text density can be a valuable characteristic in distinguishing different document layout components. For example, a paragraph usually contains more compact texts than a table with text in a sparser distribution. We construct two density-aspect graphs $\mathcal{G}_{den1}$ and $\mathcal{G}_{den2}$ to encode the text density features of each layout component based on the text density ratio and the absolute character numbers respectively.

We calculate the text density ratio for each layout component as the division between the number of character-level tokens it contains and the area size of its bounding box as in Equation~\ref{ratio}:
\begin{equation}\label{ratio}
    \text{Ratio}_{density} = \frac{\text{\#tokens}}{\text{Area size of bbox}}
\end{equation}
Given a bounding box $bbox_n$ of a layout component with the top left coordinates $\left ( x_1^{n}, y_1^{n} \right )$ and the bottom right coordinates $\left ( x_2^{n}, y_2^{n} \right )$, the area size is calculated as $\left ( x_2^{n}-x_1^{n} \right )\times \left ( y_2^{n}-y_1^{n} \right )$. We use the positional encoding approach~\citep{vaswani2017attention} to project the density ratio and the absolute character numbers of each layout component into a higher dimension $d_0=768$ to get the initialized node embedding for each density graph. The initial node embeddings of all $N$ layout components in the document page form the initial node embedding matrices of $\mathcal{G}_{den1}$ and $\mathcal{G}_{den2}$. As per Equation~\ref{pos}, $t$ denotes the value of density ratio or the absolute character numbers of each node, and $i$ is the $i$th dimension of $\overrightarrow{f_n}$. The value at $i$th dimension would change along with the odevity of $i$ for any offset $k$. 
\begin{equation}\label{pos}
   \overrightarrow{f_{n}^{i}} =  \begin{cases}
         sin(\frac{t}{10000^{2k/d_{S}}}), i=2k \\
         cos(\frac{t}{10000^{2k/d_{S}}}), i=2k+1
    \end{cases}
\end{equation}

We connect each node with its closest two neighbors with the smallest gap distance between their corresponding boxes. The edge weight is set to be the inverse distance value to emphasize the positional relationship between closer nodes. For a node $v_n$ with bounding box $bbox_n$ ($\left [ \left ( x_1^{n}, y_1^{n} \right ), \left ( x_2^{n}, y_2^{n} \right )  \right ]$), we calculate its vertical distance values with other bounding box $bbox_m$  ($\left [ \left ( x_1^{m}, y_1^{m} \right ), \left ( x_2^{m}, y_2^{m} \right )  \right ]$) that is vertically under it by $D_{V} = \left |y_{1}^{m}-y_{2}^{n}  \right |$ or $D_{V} = \left |y_{2}^{m}-y_{1}^{n}  \right |$ for bounding box that is vertically above it. This yields a set of distance values $\left \{ D_{V}^{1},...,D_{V}^{m} \right \}$, we connect the node $v_n$ with other two nodes that have the two smallest values $D_{V}$. For a two-column PDF page, in addition to the set of $D_{V}$, we also calculate the horizontal distance value $D_{H} = \left |x_{1}^{j}-x_{2}^{n}  \right |$ between $bbox_n$ and a horizontally aligned bounding box $bbox_j$ with coordinates $\left [ \left ( x_1^{j}, y_1^{j} \right ), \left ( x_2^{j}, y_2^{j} \right )  \right ]$ that has the smallest vertical gap distance. We then connect the node $v_n$ with the other two nodes with the smallest distance values among the set of $D_{V}$ and $D_{H}$. 

\subsubsection{Appearance-aspect Graph}
To learn and encode the appearance properties, such as the color and font size, of layout components in a PDF page, we use a pretrained Faster-RCNN model to extract the appearance feature $\overrightarrow{f_{appr}}$ based on the bounding box of layout component as the initial embedding for each node in $\mathcal{G}_{appr}$. We apply the same method of the edge connection and edge value setup as that for the density-aspect graphs in Section~\ref{subsec:density}.

\subsubsection{Syntactic-aspect Graph\footnote{\label{sem_syn}Details of syntactic and semantic graph generation can be found in Appendix A.2}}\label{subsec:syntactic}
To comprehensively encode the syntactic features, we use the constituency parser~\citep{kitaev2018constituency} to extract both the first-level and second-level syntactic parse of the texts in each layout component and construct two variants of the syntactic-aspect graphs: $\mathcal{G}_{syn1}$ and $\mathcal{G}_{syn2}$ respectively. The first-level parse for each layout component contains only one syntactic symbol, while the second-level parse could be a sequence of different syntactic symbols. For example, for an article title \textit{`Vitrification preserves chromatin integrity, bioenergy potential and oxidative parameters in mouse embryos'}, the first-level parse is \textit{(S)} and the second-level parse is a sequence of \textit{(NP, VP)}. We use the same positional encoding as in Equation~\ref{pos} to project a 768-dim vector $\overrightarrow{S_{w}}$ for each syntactic symbol based on their unique indexes, resulting in a sequence of syntactic embeddings $\overrightarrow{S_{1}}, \overrightarrow{S_{2}},...,\overrightarrow{S_{w}}$ for the sequence of parse for each layout component. We then feed this syntactic embedding sequence to a single-layer Bi-LSTM, and extract the last hidden state as the initialized node embedding for $\mathcal{G}_{syn2}$. Since there is only one syntactic symbol for the texts in each layout component for the first-level parse, we pad the single symbol embedding to the sequence of length $T$ for Bi-LSTM input, and get the initialized node embeddings for $\mathcal{G}_{syn1}$. For both $\mathcal{G}_{syn1}$ and $\mathcal{G}_{syn2}$, we connect every two nodes in the graph and set binary edge value $\left \{0, 1  \right \}$ based on the existence of a parent-child relationship between every two layout components. 

For the training and validation, we use the parent-child relations extracted from the document source files provided in the datasets. We apply the OCR detection to the cropped image of each document layout component. Based on the detected OCR tokens and texts in document source files, we use fuzzy string matching and a reading order assigning method proposed by \cite{ding2022v} to map each layout component with the corresponding element in the \textit{XML}/ \LaTeX\ source files for PubLayNet/DocBank and identify the parent-child relations based on the hierarchical structure embodied in these source files. We then train a transformer-based relation prediction model on the training and validation set utilizing those extracted parent-child relations and predict the parent-child relations for the test set \footnote{Details of the parent-child relation extraction and transformer-based relation prediction model are provided in Appendix A.2 and Appendix A.3}.

\subsubsection{Semantic-aspect Graph}\label{sec:semantic-aspect graph}
We use the pretrained BERT model~\citep{devlin2019bert} to encode the semantic features of each layout component to construct $\mathcal{G}_{semc}$ and extract the hidden state of the special token $[CLS]$, $\overrightarrow{f_{semc}}$, as the initial node embedding of $\mathcal{G}_{semc}$. We apply the same method for the edge connection and edge value setup as for the syntactic-aspect graphs in Section~\ref{subsec:syntactic}.

\subsection{Graph Embedding Learning}\label{sec:GCN_learning}
After the graph construction, we apply the GCN learning on each graph to update the node representations that preserve the four aspects of the layout components' properties by learning and integrating information from the neighbor nodes. For each graph $ \mathcal{G}_{i}^{p}, i \in$ \{$den1, den2, appr, syn1, syn2, semc$\} where $p$ denotes the individual PDF page of the dataset, there is an associated initial node embedding matrix $\mathcal{H}_{i}^0 \in \mathbb{R}^{N\times d_0}$, where $i \in \left \{ den1, den2, appr, syn1, syn2, semc \right \}$ as defined in Section~\ref{sec: graph_construction}. 

We feed these initialized node embedding matrices to GCN and update the weight matrices by optimizing the category prediction of each component node following the propagation rules as in Equation~\ref{gcn}. After the training, we again feed each graph $ \mathcal{G}_{i}^{p}$ into the trained GCN and extract the hidden layer node representations out, resulting in an updated node embedding matrix $ \mathcal{O}_{i}^{p} \in \mathcal{R}^{N\times d}$ for each graph $ \mathcal{G}_{i}^{p}$, where $N$ denotes the number of components in this PDF page and $d$ is the dimension of each updated node embedding.

\subsection{Multi-aspect Classification}\label{sec:3.4}
To synthesize the six graphs to four graphs with each representing one aspect of the layout components, we apply a node-level max pooling $P_m$ over the updated node embedding matrix $ \mathcal{O}_{i}^{p} \in \mathcal{R}^{N\times d}$ of the graph variants for each type of the graphs, yielding the new node embedding matrix $\mathcal{O}_{a}^{p}$ of each aspect $a$, where $a \in \left \{ density, appearance, syntactic, semantic \right \}$.

\begin{equation}
\small
    \mathcal{O}_{a}^{p} = \begin{cases} \label{max_pooling}
    P_{m}(O_{syn1}^{p}, O_{syn2}^{p}), a=syntactic\\
    P_{m}(O_{semc}^{p}, FC(\mathcal{H}_{semc}^{0,p}\prime)), a=semantic \\
    P_{m}(O_{den1}^{p}, O_{den2}^{p}), a=density \\
    P_{m}(O_{appr}^{p}, FC(\mathcal{H}_{appr}^{0,p})), a=appearance
    \end{cases}
\end{equation}

Specifically, as per Equation~\ref{max_pooling}, for syntactic-aspect and density-aspect graphs, we conduct the max pooling over the learned node representations $O_{syn1}^{p}$ ($O_{den1}^{p}$) and $O_{syn2}^{p}$ ($O_{den2}^{p}$). For semantic-aspect and appearance-aspect graphs that consists of only one graph variant, we apply the node-level max pooling over the learned node embedding matrix $O_{appr}^{p}$ and $O_{semc}^{p}$ respectively with the initial node embedding matrix $\mathcal{H}_{appr}^{0,p}$ and $\mathcal{H}_{semc}^{0,p}\prime$ that contains the fine-tuned $\overrightarrow{f_{semc}}$, where $\mathcal{H}_{appr}^{0,p}  \in \mathbb{R}^{N\times d_0}$ and $\mathcal{H}_{semc}^{0,p}\prime  \in \mathbb{R}^{N\times d_0}$ are projected into a $d$-dimensional features via fully connected layer $FC$ first. We then concatenate each of the ultimate aspects features $ \mathcal{O}_{a}^{p}$ and feed it to the 2-layer MLPs for the final classification of each document layout component in each PDF page. Our classifier is optimized based on the standard $CrossEntropy$.

\section{Evaluation Setup}

\subsection{Datasets}
We conducted our experiments on three publicly available document layout analysis datasets: PubLayNet~\citep{zhong2019publaynet}, FUNSD~\citep{jaume2019funsd} and DocBank~\citep{li2020docbank} to evaluate our model applicability. We adopted the same train/val/test split as the original dataset.\footnote{The ratio of official data split can be found in each dataset description (train/val/test)}  

\textbf{PubLayNet} annotates five different categories of layout components: \textit{Text, Title, List, Table} and \textit{Figure}, for the 358,353 PDF document images in total (94\%/3\%/3\%) collected from the PubMed.  

\textbf{FUNSD} is a much smaller dataset extracted from the RVL-CDIP dataset~\cite{harley2015evaluation} and contains 199 scanned PDF pages  (75\%/0/25\%) of survey forms with only 4 different types of layout components: \textit{Header, Question, Answer} and \textit{Other}.

\textbf{DocBank} has more sophisticated annotations for layout components than PubLayNet. It contains 500K PDF Document pages in total (80\%/10\%/10\%)\footnote{DocBank only provides the data split ratio and the entire dataset. We split the entire dataset using the same split ratio.} with 13 different categories: \textit{Abstract, Author, Caption, Date, Equation, Figure, Footer, List, Paragraph, Reference, Section, Table} and \textit{Title}. The PDF files of DocBank are collected from arXiv.com with their \LaTeX{} source.

Both \textbf{PubLayNet} and \textbf{FUNSD} include document page images, so it requires OCR\footnote{We applied Google Vision API to PubLayNet dataset, and directly used OCR result from FUNSD dataset} in order to extract the texts for each layout component. We consider those two datasets as Image-based DLA. However, \textbf{DocBank} datasets contain PDF files with text source, which contains the text for each document, so we consider this as PDF file-based DLA. We test our Doc-GCN model with both Image-based and PDF-based DLA datasets.

\subsection{Baselines}
We compare our model with four widely-used Document Layout Analysis baselines.

\textbf{Faster-RCNN}~\citep{ren2015faster} is an object detection model that unifies the region proposal network and the Fast R-CNN to extract the visual features from the proposed object regions for image object classification. For the FUNSD and DocBank, we fine-tuned the Faster-RCNN pretrained on the ImageNet with our training set and evaluated the test results. For the PubLayNet, we directly apply their Pretrained Faster-RCNN \footnote{PubLayNet Faster-RCNN: \url{https://github.com/ibm-aur-nlp/PubLayNet/}}.
    
\textbf{BERT}~\cite{devlin2019bert} is a language model which regards the DLA task as a text classification task by predicting the category of each layout component based on their sequence of text contents. 
    
\textbf{RoBERTa}~\citep{liu2019roberta} has the same structure as BERT but is pretrained longer on the larger corpus that contains longer sequences. RoBERTa also applies a more dynamic masking pattern for masked language model tasks for pre-training.
    
\textbf{LayoutLM}~\citep{xu2020layoutlm} is a pretrained model that uses the BERT architecture to jointly learn visual aspects (position) and textual features of document layouts.
\begin{table*}[t]
    \begin{center}
    \begin{adjustbox}{max width=\textwidth}
    \begin{tabular}{|p{12mm}|p{10mm}|p{8mm}|ccccc|}
    \hline
        & \textbf{\#} & \textbf{\%} & \textbf{Faster-RCNN} & \textbf{BERT-base} & \textbf{RoBERTa-base} & \textbf{LayoutLM-base} & \textbf{Ours (Doc-GCN)}\\
         \hline
        Text & 91024 & 72.87 & 96.82 & 97.74 & 97.90 & 97.94 & \textbf{99.18} \\
        Title & 19343 & 15.48 & 92.57 & 96.23 & 96.21 & 96.44 & \textbf{98.11} \\
        List & 4913 & 3.93 & 49.51 & 76.73 & 79.08 &  76.21 & \textbf{88.11}\\
        Table  & 5018 & 4.02 & 95.49 & 91.61 & 91.30 & 91.30 & \textbf{98.16}\\
        Figure & 4619 & 3.70 & 96.87 & 77.10 & 76.01 & 75.17 & \textbf{98.71}\\
        \hline
        Overall & 124917& 100 & 96.96 & 95.96 & 96.08 & 96.03 & \textbf{98.63}\\
        \hline
    \end{tabular}
    \end{adjustbox}
    \end{center}
     \caption{F1 comparison for each component category of \textbf{PubLayNet} test set. The number(\#) and percentage (\%) of each component is listed. The best models are bolded, and Doc-GCN always achieved the best for each component.}
    \label{tab:PubLayNet_breakdown}
\end{table*}

\begin{table*}[t]
    \begin{center}
    \begin{adjustbox}{max width=\textwidth}
    \begin{tabular}{|p{14mm}|p{8mm}|p{7mm}|ccccc|}
    \hline
         & \textbf{\#} &\textbf{\%}  & \textbf{Faster-RCNN} & \textbf{BERT-base} & \textbf{RoBERTa-base}  & \textbf{LayoutLM-base} & \textbf{Ours (Doc-GCN)}\\
         \hline
        Question & 1077 & 46.18 & 74.40  & 87.23 & 87.02 & 87.08 & \textbf{89.32} \\
        Answer  & 821 & 35.21 & 69.26  & 82.44 & 84.11 & 81.72 & \textbf{88.81} \\
        Header & 122 & 5.23  & 50.57 & 38.89 & 45.98 & 44.90 & \textbf{61.96}\\
        Other & 312 & 13.38 & 65.12 & 57.33 & 56.20 & 56.19 & \textbf{72.76} \\
        \hline
        Overall & 2332 & 100 & 65.12 & 79.02 & 81.69 & 78.85 & \textbf{85.49}\\
        \hline
    \end{tabular}
    \end{adjustbox}
    \end{center}
    \caption{F1 comparison for each component of \textbf{FUNSD} test set. The number(\#) and percentage (\%) of each component is listed. The best models are bolded, and Doc-GCN always achieved the best for each component.}
    \label{tab:FUNSD_breakdown}
\end{table*}

\begin{table*}[ht!]
    \begin{center}
    \begin{adjustbox}{max width=\textwidth}
    \begin{tabular}{|p{14mm}|p{8mm}|p{7mm}|ccccc|}
    \hline
         & \textbf{\#} & \textbf{\%} & \textbf{Faster-RCNN} & \textbf{BERT-base} & \textbf{RoBERTa-base} & \textbf{LayoutLM-base} & \textbf{Ours (Doc-GCN)} \\
         \hline
        Abstract & 420 & 0.70 & 0 & 70.04 & 60.61 & 58.27 & \textbf{78.69} \\
        Author & 484 & 0.80 & 0 & 72.56 & \textbf{82.01} & 69.07 & 79.46 \\
        Caption & 1840 & 3.05 & 4.44 & 77.33 & 76.48 & 74.78 & \textbf{87.38} \\
        Date & 87 & 0.14 & 0 & 67.76 & 88.74 & 85.35 & \textbf{91.02} \\
        Equation & 11846 & 19.66 & 81.80 & 85.92 & 86.08 & 86.00 & \textbf{90.06} \\
        Figure & 1650 & 2.74 & 68.74 & \textbf{100} & \textbf{100} & \textbf{100} & 99.97 \\
        Footer & 529 & 0.88 & 0 & 69.08 & 68.67 & 65.56 & \textbf{84.48} \\
        List & 958 & 1.59 & 0 & 54.72 & 55.01 & 50.43 & \textbf{65.79} \\
        Paragraph & 35496 & 58.92 & 83.69 & 89.44 & 89.69 & 89.05 & \textbf{96.50} \\
        Reference & 1237 & 2.05 & 0 & 88.51 & 88.30 & \textbf{89.61} & 88.83 \\
        Section & 4891 & 8.12 & 78.52 & 84.04 & 84.79 & 83.90 & \textbf{95.93} \\
        Table & 525 & 0.87 & 0 & 49.29 & 51.79 & 49.17 & \textbf{59.06} \\
        Title & 286 & 0.47 & 0 & 51.49 & 68.58 & 55.53 & \textbf{84.85} \\
        \hline
        Overall & 60249 & 100 & 71.02 & 86.65 & 86.97 & 86.16 & \textbf{91.83} \\
        \hline
    \end{tabular}
    \end{adjustbox}
    \end{center}
    \caption{F1 comparison for each component of \textbf{DocBank} test set. The number(\#) and percentage (\%) of each component is listed. The best models are bolded, and Doc-GCN always achieved the best/the second best.}
    \label{tab:DocBank_breakdown}
\end{table*}

\begin{table*}[t]
    \begin{center}
    \begin{adjustbox}{max width=\textwidth}
    \begin{tabular}{|cccc|ccc|ccc|ccc|}
        \hline
         & & & & \multicolumn{3}{|c|}{\textbf{PubLayNet}} & \multicolumn{3}{|c|}{\textbf{FUNSD}} & \multicolumn{3}{|c|}{\textbf{DocBank}}\\
        \hline
         \textbf{\textit{Syn}} & \textbf{\textit{Sem}} & \textbf{\textit{Dens}} & \textbf{\textit{Appr}} & Precision & Recall & F1 & Precision & Recall & F1 & Precision & Recall & F1\\
         \hline
         O & X & X & X  & 67.81 & 72.92 & 63.45 & 54.80 & 49.36 & 39.68 & 40.95 & 46.01 & 42.40\\
         X & O & X & X  & 96.70 & 96.37 & 96.48 & 81.68 & 82.33 & 81.84 & 88.68 & 88.54 & 88.61\\
         X & X & O & X  & 50.65 & 71.17 & 59.18 & 45.91 & 48.93 & 43.52 & 34.71 & 58.90 & 43.68\\
         X & X & X & O  & 96.53 & 95.95 & 96.15 & 70.82 & 70.88 & 70.63 & 87.18 & 85.91 & 86.27\\
         X & O & X & O  & 98.42 & 98.36 & 98.38 & 84.45 & \underline{84.99} & 84.41 & 91.80 & \underline{91.79} & 91.66\\
         X & O & O & O  & 98.49 & 98.45 & 98.46 & 84.55 & 84.31 & 84.42 & 91.95 & \textbf{91.94} & \underline{91.74}\\
         O & O & X & O  & \underline{98.61} & \underline{98.58} & \underline{98.59} & \underline{84.85} & \underline{84.99} & \underline{84.85} & \underline{92.00} & 91.77 & 91.68\\
         \hline
          O & O & O & O  &\textbf{98.64} & \textbf{98.65} &\textbf{98.63} & \textbf{85.54} & \textbf{85.55} & \textbf{85.49} & \textbf{92.07} & \textbf{91.94} &\textbf{91.83}\\
        \hline
    \end{tabular}
    \end{adjustbox}
    \end{center}
    \caption{F1 comparison of using different aspects of graph features on the validation set of three datasets. \textit{Syn}, \textit{Sem}, \textit{Dens} and \textit{Appr} stand for the Syntactic-based, Semantic-based, Density-based and Appearance-based graphs respectively. "O" and "X" refer to the existence and absence of corresponding graph features for the classification. The second best is \underline{underlined}.}
    \label{tab:ablation}
\end{table*}

\subsection{Implementation Details}
For node features of Semantic graph and Appearance graph, we extract $[CLS]$ encoding ($dim=768$) from pre-trained $BERT_{BASE}$. The Faster-RCNN with ResNet-101 pre-trained on ImageNet is used to extract the visual features ($dim=2048$) to initialize the node embedding of the appearance graph. We set $T = 16$ for the sequence padding in syntactic graph construction. All the graphs are trained on 2-layer GCNs for 10 epochs with Adam optimizer using a learning rate of $1\times 10^{-4}$ for Semantic and Syntactic Graph and a learning rate of 0.001 for the other two graphs. We use $ReLu$ as the activation function for the first GCN layer, from which we extract our learned node representations. For final classification, we train the multi-aspect classifier with Adam optimizer using the learning rate of $2\times 10^{-5}$, dropout of 0.1 and $Tanh$ as activation function. We trained the model using Intel(R) Xeon(R) CPU @ 2.00GHz and NVIDIA Tesla K80 24GB, which took around 15 hours, 72 hours and 8 mins\footnote{With more aspects covered, Doc-GCN training is still slightly smaller than other baselines recorded in each paper} for training PubLayNet, DocBank and FUNSD respectively. The number of trainable parameters is 126,401,796.

\section{Results}
\subsection{Performance Comparison}
We compared the performance of our proposed Doc-GCN with the baseline models on the test set of PubLayNet, FUNSD, and DocBank. The breakdown F1 scores on each layout component of the three datasets are provided in Table~\ref{tab:PubLayNet_breakdown},~\ref{tab:FUNSD_breakdown} and~\ref{tab:DocBank_breakdown} respectively. 

We can see from the overall F1 scores that Doc-GCN outperforms all the baseline models on all datasets. It achieves a 98.63\% F1 score in PubLayNet, which is 1.67\% higher than the pre-trained Faster-RCNN. The performances of Faster-RCNN on FUNSD and DocBank are also much less competitive than models utilizing semantic features, which indicates the critical role of semantic information in the DLA task. Compared to BERT and RoBERTa, which only uses the semantic features, the F1 scores of our model are 2.55\%, 3.8\%, 4.86\% higher than RoBERTa and 2.67\%, 6.47\%, 5.18\% higher than BERT on PubLayNet, FUNSD and DocBank respectively. Especially for FUNSD, where there are only 144 training PDF pages, our model still achieves around 85\%. Such results also prove the effectiveness of our multiple aspects of document layout component features for low-resource DLA datasets. Though LayoutLM also utilizes dual-aspect, visual aspect (position), and textual features, our Doc-GCN produces better overall F1 scores on all datasets. The superior performance of Doc-GCN demonstrates the effectiveness of the multiple aspects we applied. It indicates the positive contribution of graph representations and the joint GCN learning for integrating multiple features in the DLA task.

Furthermore, Doc-GCN shows significant performance improvement for some important layout component detection for each domain. It produce a better result in all components in PubLayNet, including the \textit{Title}, \textit{Table}, \textit{Figure}, and the \textit{List} that contains the structural text information, as is demonstrated in Table~\ref{tab:PubLayNet_breakdown}. Similar patterns can be also observed in Table~\ref{tab:FUNSD_breakdown} for FUNSD, especially with those important components of the forms: \textit{Header} (section title - e.g. the aim of the form) and \textit{Answer} (value - e.g. the content filled by the form user). For DocBank in Table~\ref{tab:DocBank_breakdown}, it shows that Doc-GCN works well with the essential components of scientific academic publication, including \textit{Abstract}, \textit{Date}, \textit{Section}, \textit{Title}. Those components are generally extracted for understanding the content of documents.


\subsection{Comparison of Aspect Variants}
To inspect the effectiveness of each aspect feature captured by Doc-GCN, we further compare its test performances by applying different combinations of aspect features. As seen in Table~\ref{tab:ablation}, using only semantic or appearance features already resulted in around 96\% and 88\% in F1 scores on PubLayNet and DocBank, respectively. For FUNSD, using only appearance features results in 70.63\% in the F1 score, and using only semantic features is around 10\% higher. The combination of semantic and appearance features improves the F1 scores to 98.38\%, 84.41\%, and 91.66\% for PubLayNet, FUNSD, and DocBank, respectively. The semantic and appearance features seem to dominate the model's performance in the DLA task, but the syntactic and density features also positively contribute to the performance. By adding syntactic features to the semantic and appearance features, the F1 scores on PubLayNet and FUNSD improve to 98.59\% and 84.85\%, respectively. Further, adding density features results in the best performances of 98.63\% and 85.49\%, respectively. For DocBank, adding density features improves the F1 score from 91.66\% to 91.74\% and finally reaches the best performance of 91.83\% after further including the syntactic features. Though the density features and syntactic features contribute differently in the cases of different datasets, it is evident that the utilization of both density features and syntactic features is effective for performance improvement in the DLA task. Such results also indicate that our proposed aspects and their representations are practical for a more comprehensive representation of the characteristics of document layout components.

\begin{table}[t]
    \begin{center}
    \begin{adjustbox}{max width = \linewidth}
    \begin{tabular}{|l|ccc|}
        \hline
         & \textbf{PubLayNet} & \textbf{FUNSD} & \textbf{DocBank}\\
         \hline
         \textbf{Bert-L} & 97.06 & 80.86 & 86.42\\
         \textbf{RoBERTa-L} & 96.15 & 79.47 & 86.67\\
         \textbf{LayoutLM-L} & 96.94 & 78.90 & 83.21\\
         \hline
         \textbf{Doc-GCN-L} & \textbf{98.22} & \textbf{85.40} & \textbf{90.07}\\
         \hline
    \end{tabular}
    \end{adjustbox}
    \end{center}
    \caption{Performances comparison (F1 score\%) based on large (L) pretrained models.}
    \label{tab:base_large}
\end{table}
\begin{table}[t]
    \begin{center}
    \begin{adjustbox}{max width = \linewidth}
    \begin{tabular}{|l|ccc|}
    \hline
         & \textbf{PubLayNet} & \textbf{FUNSD} & \textbf{DocBank}\\
        \hline
         \textbf{Min Pooling} &\textbf{98.67} & 82.77 & 90.27\\
         \textbf{Avg Pooling} & 97.90 & 82.04 & 89.04\\
         \textbf{Max Pooling} &98.63 & \textbf{85.49} & \textbf{91.83}\\
         \hline
    \end{tabular}
    \end{adjustbox}
    \end{center}
    \caption{Performances comparison (F1 score\%) of different node-level pooling methods on the test set.}
    \label{tab:pooling}
\end{table}

\begin{figure*}[t]
    \hspace*{-0.5cm}
    \captionsetup[subfigure]{aboveskip=-10pt,belowskip=-4pt}
     \centering
     \begin{subfigure}[b]{0.25\textwidth}
         \centering
         \includegraphics[width=4.1cm]{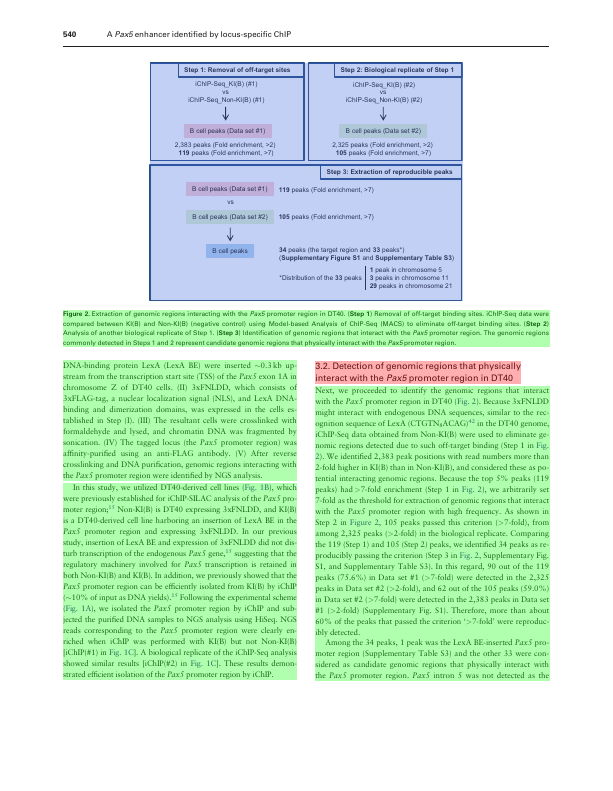}
         \caption{\textbf{Ground Truth}}
         \label{fig:case_study_GT_edite}
     \end{subfigure}
     \hspace*{-0.8em}
     \begin{subfigure}[b]{0.25\textwidth}
         \centering
         \includegraphics[width=4.1cm]{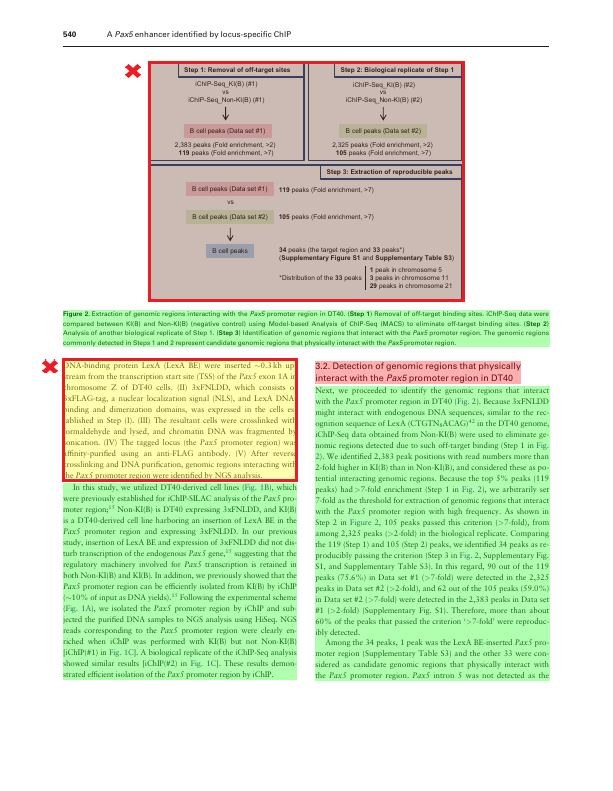}
         \caption{\textbf{RoBERTa-base}}
         \label{fig:case_study_bert}
     \end{subfigure}
     \hspace*{-0.8em}
     \begin{subfigure}[b]{0.25\textwidth}
         \centering
         \includegraphics[width=4.1cm]{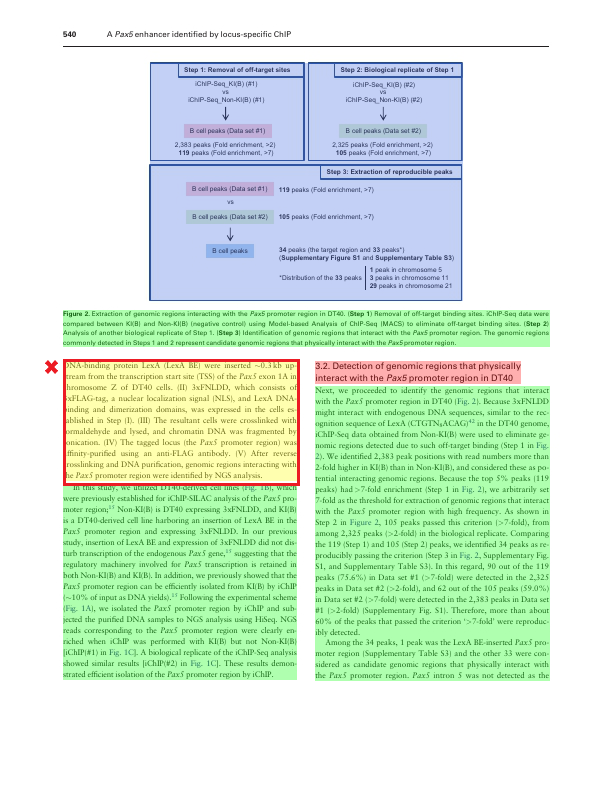}
         \caption{\textbf{Faster-RCNN}}
         \label{fig:case_study_GT}
     \end{subfigure}
     \hspace*{-0.8em}
     \begin{subfigure}[b]{0.25\textwidth}
         \centering
         \includegraphics[width=4.1cm]{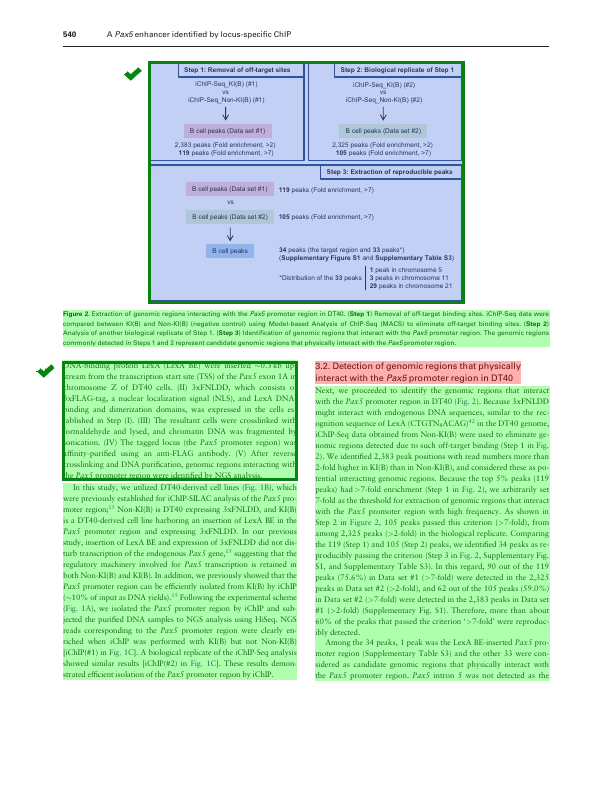}
         \caption{\textbf{Doc-GCN}}
         \label{fig:five over x}
     \end{subfigure}
        \caption{Example output of Top3 models for a PubLayNet page. The color of layout component labels are: \colorbox{text}{Text}, \colorbox{title}{Title}, \colorbox{list}{List}, \colorbox{table}{Table}, \colorbox{figure}{Figure}. Our Doc-GCN classified all layout components accurately.}
        \label{fig:sample_output}
\end{figure*}

\subsection{Impact of Pretrained Model Variants}
To evaluate the effects of base and large pretrained models, we also tested the performances of large BERT, RoBERTa and LayoutLM over three datasets. For a fair comparison, we deployed Doc-GCN Large, shown as Doc-GCN-L in Table~\ref{tab:base_large}, which uses the same graph construction and representations as the original Doc-GCN except that the large pretrained BERT is used for generating the semantic-aspect graph features. 

The result shows that large models have similar performances as base models. Nevertheless, our Doc-GCN-L still outperforms BERT-L, RoBERTa-L and LayoutLM-L on all three datasets: PubLayNet, FUNSD, and DocBank.

\subsection{Impact of Pooling Variants}
We applied the node-level pooling to integrate the node features over the two graph variants as the final features for each aspect representation. We compared Minimum, Average, and Max Pooling and used the method with the best results. From Table~\ref{tab:pooling}, we can see that for FUNSD and DocBank, Max Pooling resulted in the best performances. Especially for the FUNSD dataset, Max Pooling achieved almost 3\% higher in F1 score compared to the results of using Minimum Pooling and Average Pooling. For PubLayNet, Minimum Pooling results in the best performance but is only 0.03\% higher than Max Pooling. Hence, we used Max Pooling as the ultimate pooling method in our Doc-GCN.

\subsection{Case Study\footnote{More qualitative analysis for each of the three datasets can be found in Appendix B, and the superiority of understanding multiple aspects will be highlighted there.}}

We visualized the sample results for the top 3 models on a document page of PubLayNet in Figure~\ref{fig:sample_output}. We can see that both RoBERTa and Faster-RCNN have wrongly recognized a \textit{Text} into \textit{List}, whereas our Doc-GCN has accurately recognized all components. This case further proves that simply considering the semantic or visual information is hard to distinguish the \textit{List} and \textit{Text}, indicating the importance of capturing the structural relationships between layout components for better performance. 

\section{Conclusion\footnote{Appendix: \url{https://github.com/adlnlp/doc_gcn/tree/main/appendix}}}
We successfully handled the importance of the DLA task, Document Component/Segment Classification. It focused on extracting important information (i.e., Title, Author, Date, Form Answer) from the digital documents, including Scanned Document Images and PDF files. We propose a heterogeneous graph-based DLA model, Doc-GCN, which integrates text density, appearance, and syntactic and semantic properties of the Document layout components to generate comprehensive representations of documents. The graph structure also enables the model to capture and learn the relationships between the layout components when making predictions. Our model outperforms all baselines on three publicly available DLA datasets. We strongly hope that Doc-GCN will motivate and provide insights into the future integration of different modalities and aspects for the logical document layout analysis task.

\bibliography{anthology,custom}
\bibliographystyle{acl_natbib}
\newpage
\appendix

\section{Graph Construction Procedure}

\begin{figure*}[t]
     \centering
     \includegraphics[scale=0.53]{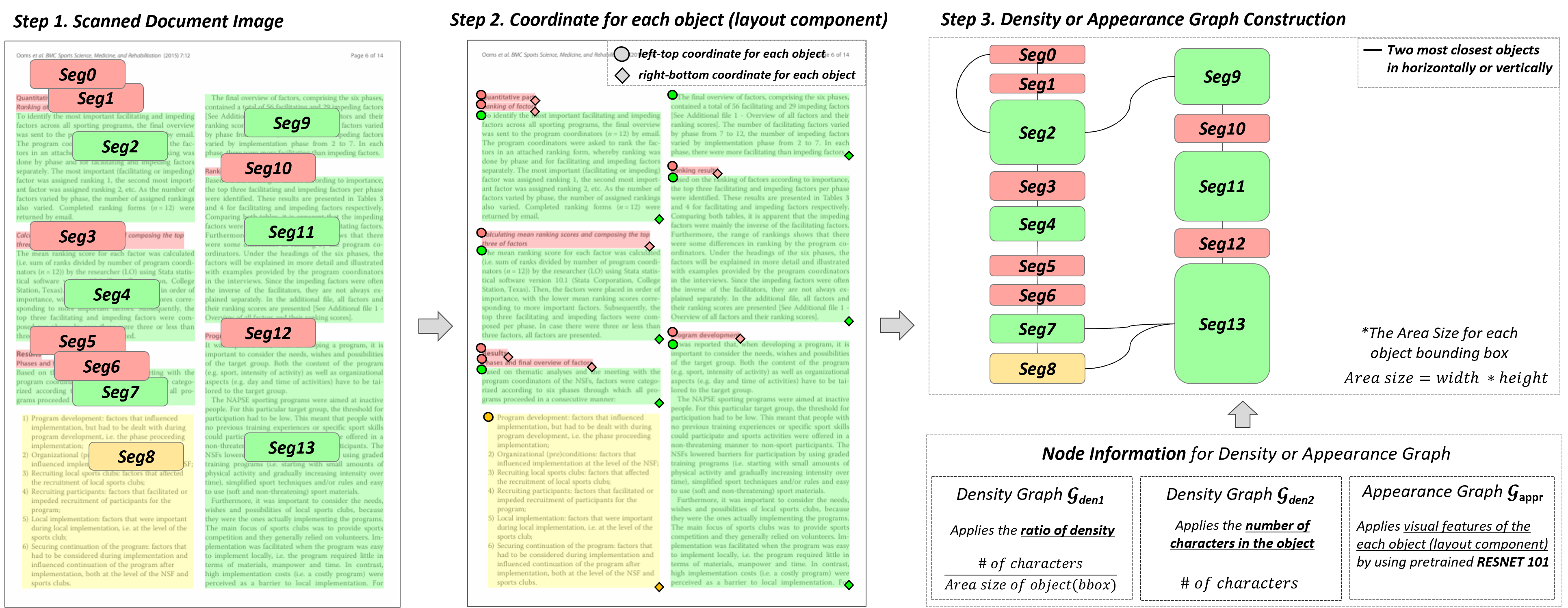}
    \caption{Density Graph and Appearance Graph Construction}
    \label{fig:density_app_drawing}
\end{figure*}

\begin{figure*}[ht]
    \hspace*{0cm}
     \centering
     \begin{subfigure}[b]{\textwidth}
         \centering
         \includegraphics[width=16cm]{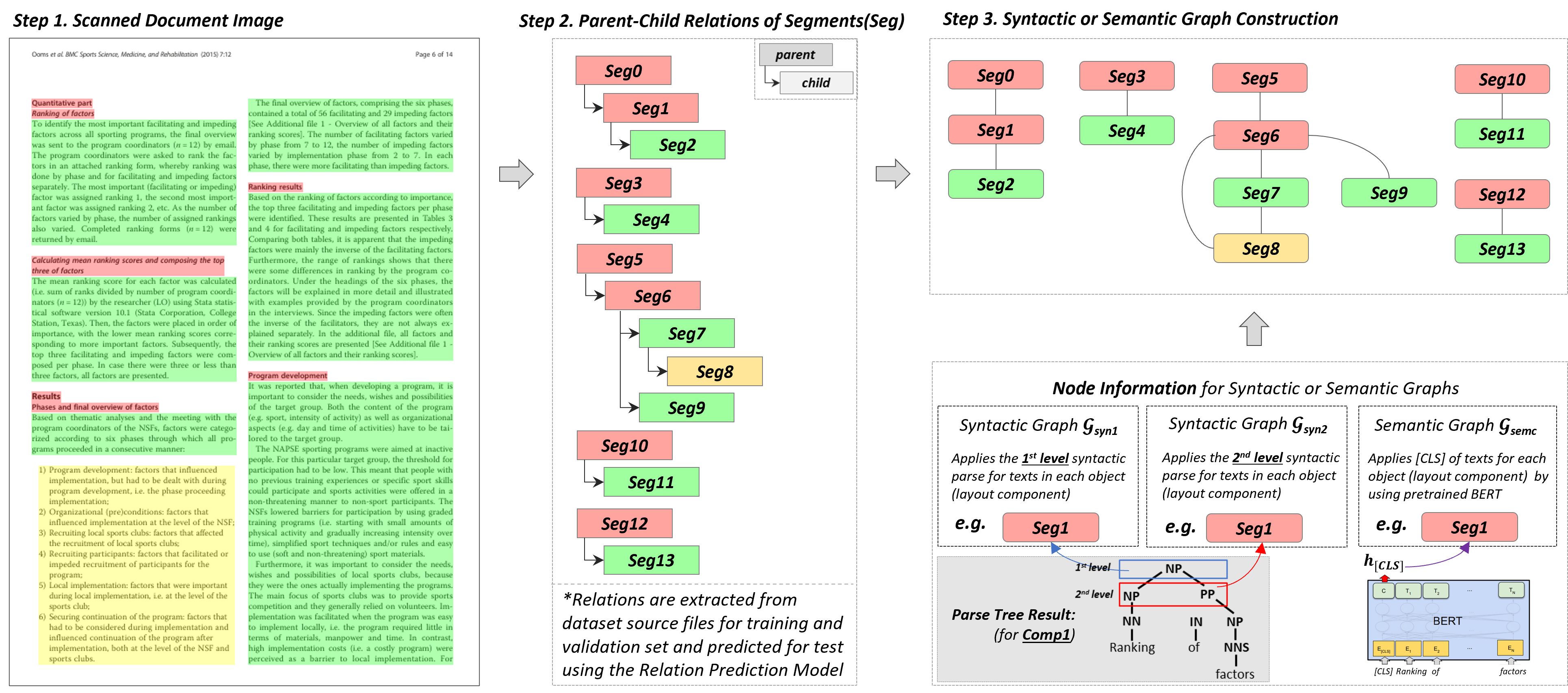}
         \caption{Semantic Graph and Syntactic Graph Construction}
         \label{fig:taskb_answer_word_cloud}
     \end{subfigure}
     \vspace*{2cm}
     \begin{subfigure}[b]{\textwidth}
         \centering
         \includegraphics[width=14cm]{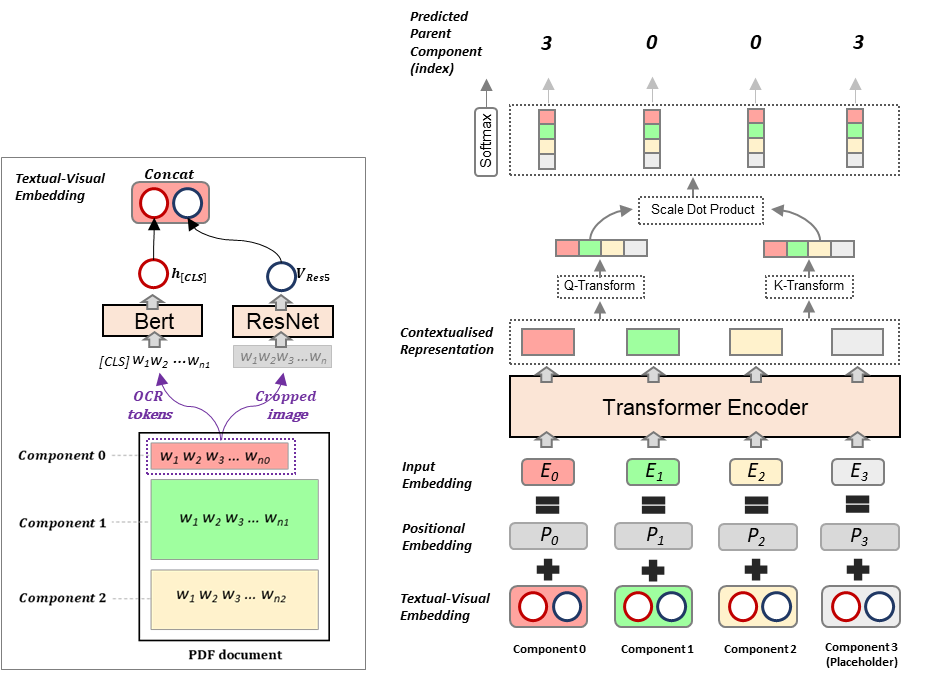}
         \caption{Relation Prediction Model}
         \label{fig:taskb_answer_word_cloud_struct}
     \end{subfigure}
        \caption{Semantic Graph and Syntactic Graph Construction flow with Parent-Child Relation prediction model architecture.}
        \label{fig:ans_wordcloud}
\end{figure*}
The proposed Doc-GCN contains graphs covering four different aspects of properties of PDF pages, including syntactic, semantic, density, and appearance/visual information, to harmonize and integrate heterogeneous aspects for Document Layout Analysis. In this section, we would like to describe the graph construction procedure for each aspect graph. The syntactic and semantic graphs share the same graph structure, while the density and appearance graphs use the other graph structures.

\subsection{Density-aspect and Appearance-aspect Graph construction}
The graph construction scenarios for both density-aspect and appearance-aspect graphs can be found in Figure~\ref{fig:density_app_drawing}. We demonstrate the detailed steps as follows: 

\begin{enumerate}
    \item As is illustrated in step 1, given a scanned document page, we identify each document layout component as a segment based on the bounding box provided in the datasets.
    
    \item In step 2, the bounding box information for each segment is defined by the left-top and right-bottom coordinates. 

    \item To construct the Density-aspect/Appearance-aspect graphs in step 3, we treat each segment as a graph node and connect each segment node with its two closest neighbors based on the gap between their bounding boxes. These two closest neighbors could be both vertically and horizontally located (if a two-column page is applied). We initialize the $\mathcal{G}_{den1}$ node values as the ratio of density following the Equation~\ref{ratio}.
    \begin{equation}\label{ratio}
    \text{Ratio}_{density} = \frac{\text{\#tokens}}{\text{Area size of bbox}}
    \end{equation}
    $\mathcal{G}_{den2}$ node values are initialized as the total number of characters in each segment.
    
    For appearance graph, $\mathcal{G}_{appr}$, the node values are initialized by the visual features of each segment. Such visual features are extracted from the pretrained ResNet-101. 
    
\end{enumerate}

\subsection{Syntactic-aspect and Semantic-aspect Graph Construction}\label{sec:syn and sem}
The graph construction scenarios for both syntactic-aspect and semantic-aspect graphs can be found in Figure~\ref{fig:taskb_answer_word_cloud}. We demonstrate the detailed steps as follows:



\begin{enumerate}
    \item As is illustrated by step 1, given a scanned document page, we identify each document layout component as a segment based on the bounding box provided in the dataset.
    
    \item In step 2, we represent the hierarchical parent-child relations between segments. For the training and validation set, we use the parent-child relations that are directly derived from the source files provided in the datasets: we first apply OCR detection to each segment in step 1. Based on the detected OCR text tokens, we then use fuzzy string matching to map each segment with the corresponding element in the \textit{XML}/ \LaTeX\ source files for PubLayNet/DocBank and identify the parent-child relations based on the hierarchical structure embodied in these source files; FUNSD provides the parent-child relations directly in the $.json$ annotation files. We then train a transformer-based relation prediction model on the (training + validation) set utilizing those extracted parent-child relations and predict the parent-child relations for the test set. The details of the transformer-based relation prediction model are provided in Section~\ref{sec:relationmodel}.
    
    \item To construct the Syntactic and Semantic graph, as is demonstrated in step 3, we treat each segment as a node of the graph. We then connect the parent-child relations nodes (derived from step 2) and set the edge value as 1. Regarding the node representation, we use the first-level and second-level syntactic parsing results as the node information for the Syntactic Graph $\mathcal{G}_{syn1}$ and $\mathcal{G}_{syn2}$ respectively before the node encoding. For the Semantic Graph ${G}_{semc}$, we pass the OCR tokens within each segment node to the pretrained BERT and extract the $[CLS]$ as the initialized node embedding.
    
    

\end{enumerate}

\subsection{Relation Prediction Model}\label{sec:relationmodel}
As described in Section~\ref{sec:syn and sem}, we apply a relation prediction model to predict the parent-child relations between segments when constructing both syntactic-aspect and semantic-aspect graphs for data in the test set. 

Given a PDF page with multiple segments, we first encode each segment's textual and visual features, as illustrated on the left side in Figure~\ref{fig:taskb_answer_word_cloud_struct}. On the one hand, we feed the OCR token sequence $w_1, w_2,...,w_n$ (derived from step 2 in Section~\ref{sec:syn and sem}) of the segment into the pretrained BERT model and adopt the generated representation of the $[CLS]$ token as the textual embedding $h_{[CLS]}$. On the other hand, the cropped image of each segment (based on the bounding box) is fed into the pretrained ResNet-101 model, from which the output feature of Res5 layer is used as the visual embedding $V_{Res5}$. We then concatenate the textual embedding $h_{[CLS]}$ and visual embedding as the textual-visual embedding for each segment.
\begin{table}[t]
    \begin{center}
    \begin{adjustbox}{max width = \linewidth}
    \begin{tabular}{|l|ccc|}
        \hline
         & \textbf{PubLayNet} & \textbf{FUNSD} & \textbf{DocBank}\\
         \hline
         \textbf{Bert-L} & 97.06 & 80.86 & 86.42\\
         \textbf{RoBERTa-L} & 96.15 & 79.47 & 86.67\\
         \textbf{LayoutLM-L} & 96.94 & 78.90 & 83.21\\
         \hline
         \textbf{Doc-GCN-L} & \textbf{98.22} & \textbf{85.40} & \textbf{90.07}\\
         \hline
    \end{tabular}
    \end{adjustbox}
    \end{center}
    \caption{Performances comparison (F1 score\%) based on large (L) pretrained models.}
    \label{tab:base_large}
\end{table}

The relation prediction model is shown on the right side in Figure~\ref{fig:taskb_answer_word_cloud_struct}. We represent the sequence of segments in a PDF page using their textual-visual embedding and sum it up with the positional embedding as the input embedding, which is then passed to a single-layer transformer encoder to derive the contextualized representation for the segments. We then apply two linear transformations, Q-Transform and K-Transform, to generate the $query$ and $key$ representation. An attention score matrix is generated from the softmax-normalized dot-product results of the two representations, based on which the parent segment is selected from the segments in the input sequence. To handle those segments without parents, we append a fake segment at the end of the input sequence initialized with all zeros as a place holder, which is treated as the target parent for the non-parent segment.

\section{Additional Case Studies}
\begin{figure*}[h]
    \hspace*{-0.5cm}
    \captionsetup[subfigure]{aboveskip=-10pt,belowskip=-4pt}
     \centering
     \begin{subfigure}[b]{0.25\textwidth}
         \centering
         \includegraphics[width=4.1cm]{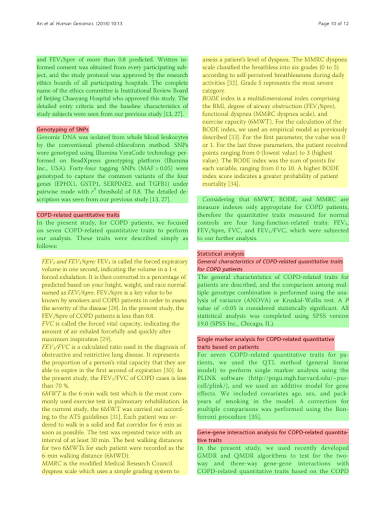}
         \caption{\textbf{Ground Truth}}
         \label{fig:GT_PubLayNet_appendix}
     \end{subfigure}
     \hspace*{-0.8em}
     \begin{subfigure}[b]{0.25\textwidth}
         \centering
         \includegraphics[width=4.1cm]{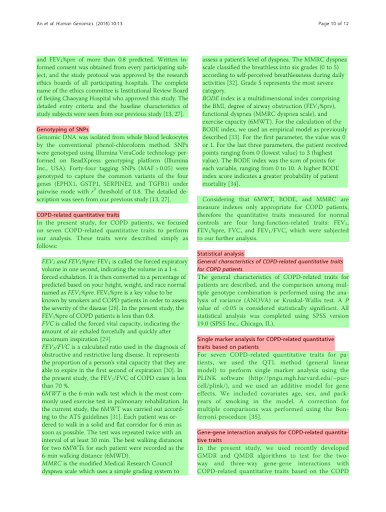}
         \caption{\textbf{RoBERTa-base}}
         \label{fig:Roberta_Publaynet_appendix}
     \end{subfigure}
     \hspace*{-0.8em}
     \begin{subfigure}[b]{0.25\textwidth}
         \centering
         \includegraphics[width=4.1cm]{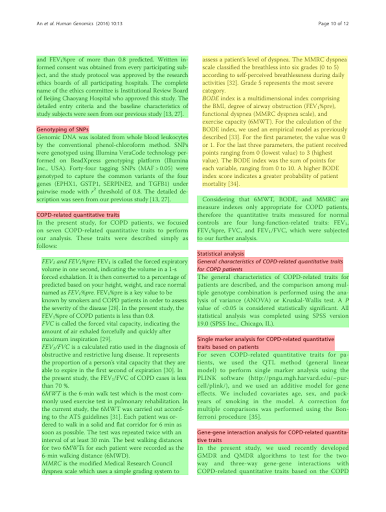}
         \caption{\textbf{Faster-RCNN}}
         \label{fig:FasterRCNN_PubLayNet_appendix}
     \end{subfigure}
     \hspace*{-0.8em}
     \begin{subfigure}[b]{0.25\textwidth}
         \centering
         \includegraphics[width=4.1cm]{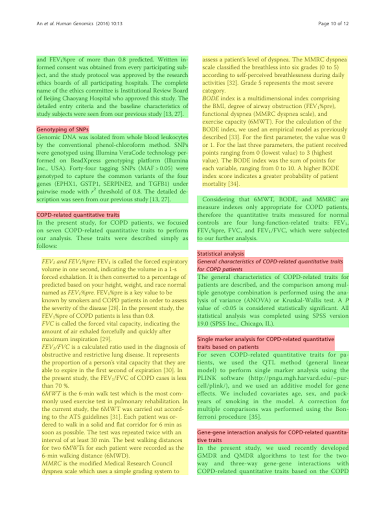}
         \caption{\textbf{Doc-GCN}}
         \label{fig:DocGCN_PubLayNet_appendix}
     \end{subfigure}
        \caption{Example output of Top3 models for a \textbf{PubLayNet} page. The color of layout component labels are: \colorbox{text}{Text}, \colorbox{title}{Title}, \colorbox{list}{List}, \colorbox{table}{Table}, \colorbox{figure}{Figure}. Our Doc-GCN classified all layout components accurately.}
        \label{fig:publaynet_addition_case}
\end{figure*}

\begin{figure*}[h]
    \hspace*{-0.6cm}
    \captionsetup[subfigure]{aboveskip=-2pt,belowskip=-2pt}
     \centering
     \begin{subfigure}[b]{0.25\textwidth}
         \centering
         \includegraphics[width=4.1cm]{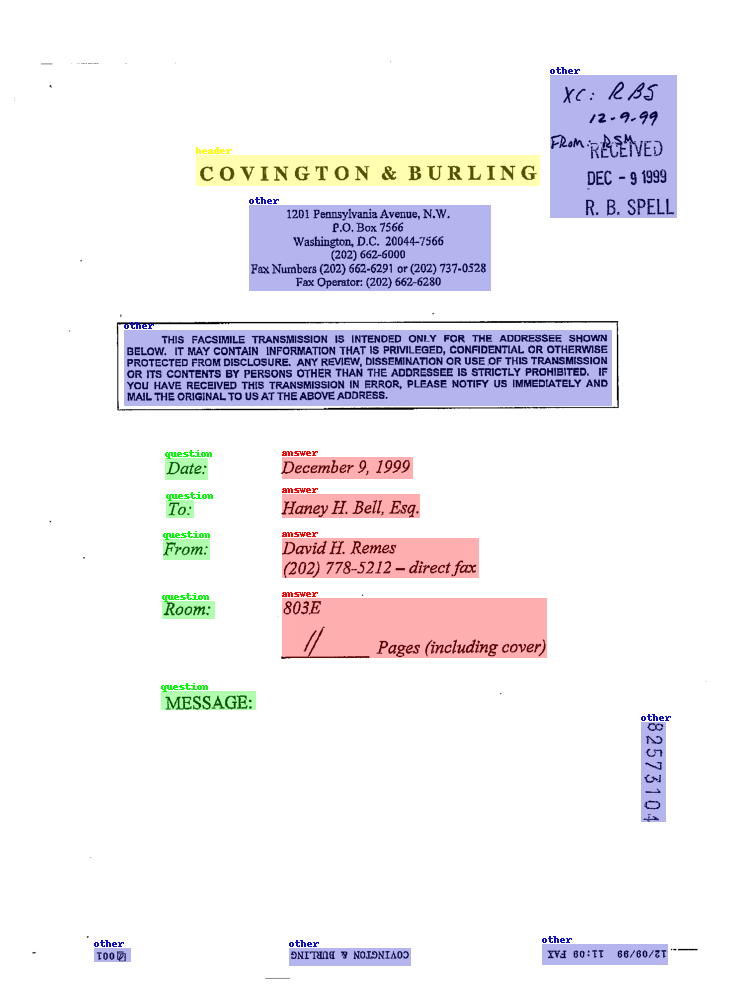}
         \caption{\textbf{Ground Truth}}
         \label{fig:case_study_GT}
     \end{subfigure}
     \hspace*{-0.8em}
     \begin{subfigure}[b]{0.25\textwidth}
         \centering
         \includegraphics[width=4.1cm]{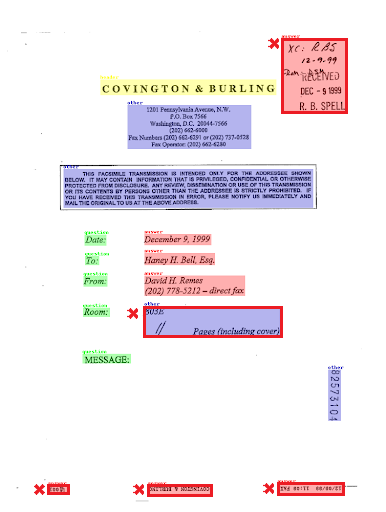}
         \caption{\textbf{BERT-base}}
         \label{fig:case_study_bert}
     \end{subfigure}
     \hspace*{-0.8em}
     \begin{subfigure}[b]{0.25\textwidth}
         \centering
         \includegraphics[width=4.1cm]{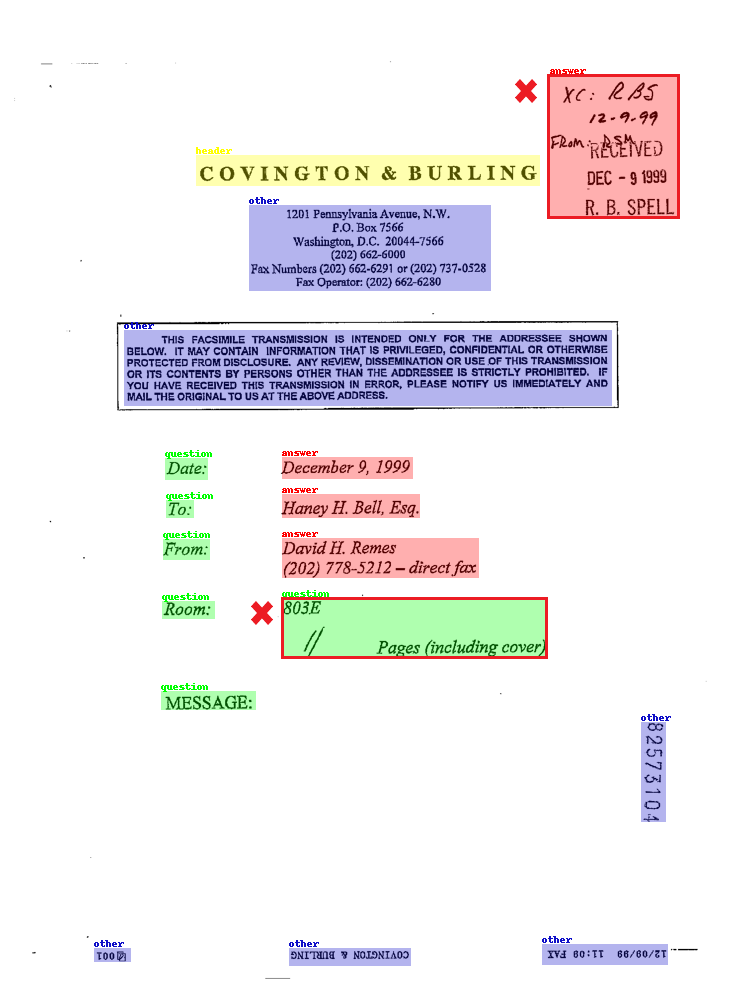}
         \caption{\textbf{RoBERTa-base}}
         \label{fig:case_study_layoutlm}
     \end{subfigure}
     \hspace*{-0.8em}
     \begin{subfigure}[b]{0.25\textwidth}
         \centering
         \includegraphics[width=4.1cm]{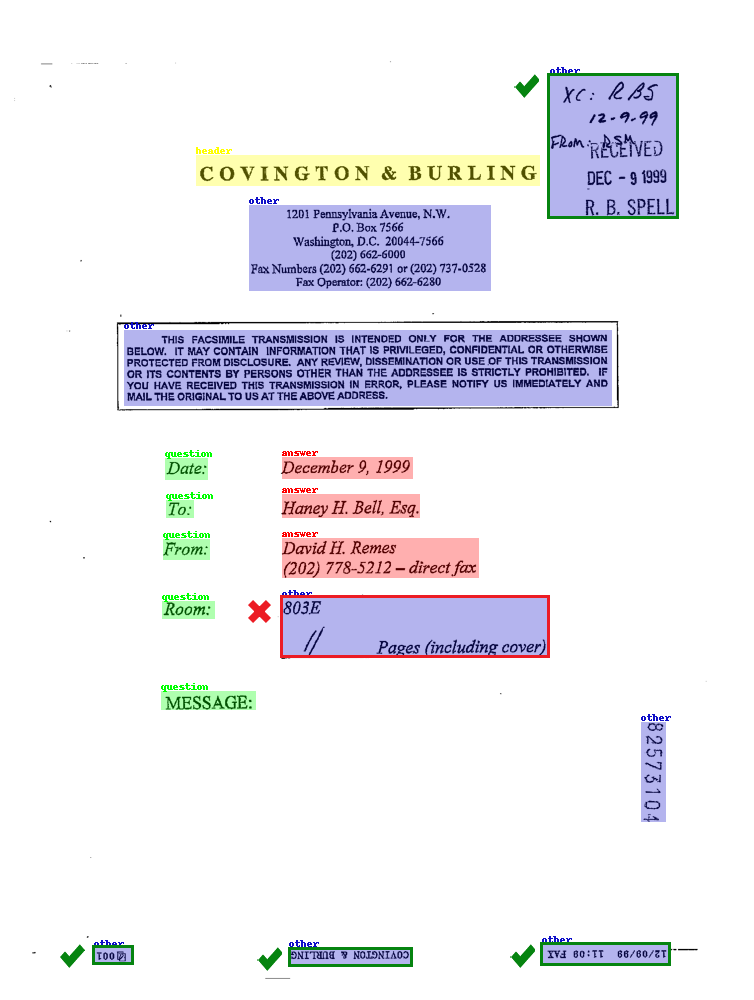}
         \caption{\textbf{Doc-GCN}}
         \label{fig:funsd_ourmodel}
     \end{subfigure}
        \caption{Visualisation of Top3 models for a FUNSD page. The color of layout components are: \colorbox{question}{Question}, \colorbox{answer}{Answer}, \colorbox{header}{Header}, \colorbox{other}{Other}. Our Doc-GCN produced the best recognition results, especially for the \textit{Other} components.}
        \label{fig:vis_funsd}
\end{figure*}

\begin{figure*}[h]
    \hspace*{-0.5cm}
    \captionsetup[subfigure]{aboveskip=-2pt,belowskip=-2pt}
     \centering
     \begin{subfigure}[b]{0.25\textwidth}
         \centering
         \includegraphics[width=4.1cm]{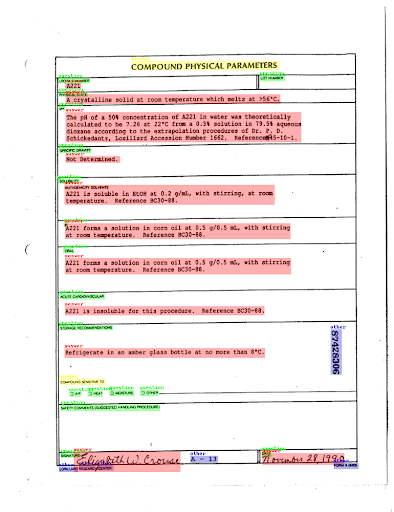}
         \caption{\textbf{Ground Truth}}
         \label{fig:GT_FUNSD_appendix}
     \end{subfigure}
     \hspace*{-0.8em}
     \begin{subfigure}[b]{0.25\textwidth}
         \centering
         \includegraphics[width=4.1cm]{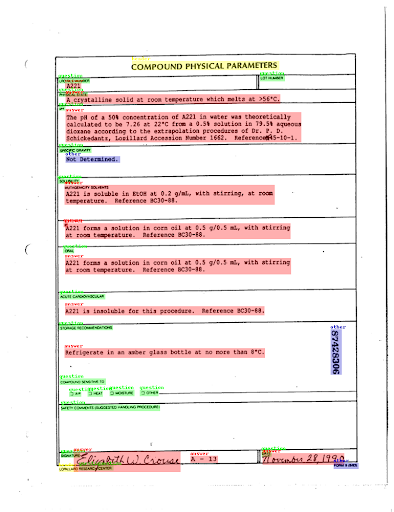}
         \caption{\textbf{BERT-base}}
         \label{fig:BERT_FUNSD_appendix}
     \end{subfigure}
     \hspace*{-0.8em}
     \begin{subfigure}[b]{0.25\textwidth}
         \centering
         \includegraphics[width=4.1cm]{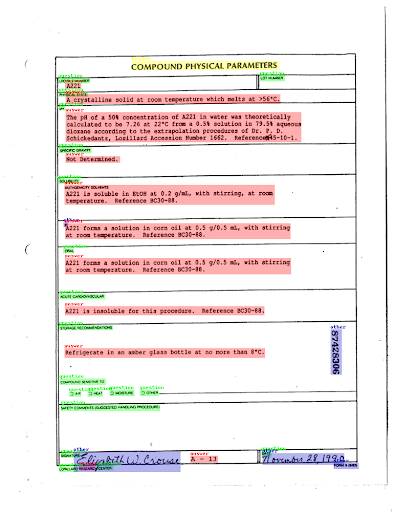}
         \caption{\textbf{RoBERTa-base}}
         \label{fig:Roberta_FUNSD_appendix}
     \end{subfigure}
     \hspace*{-0.8em}
     \begin{subfigure}[b]{0.25\textwidth}
         \centering
         \includegraphics[width=4.1cm]{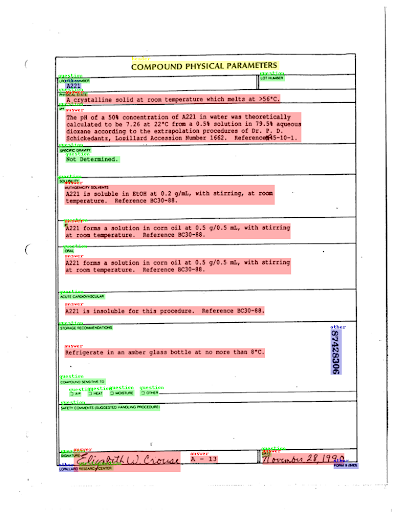}
         \caption{\textbf{Doc-GCN}}
         \label{fig:DocGCN_FUNSD_appendix}
     \end{subfigure}
        \caption{Visualisation of Top3 models for a \textbf{FUNSD} page. The color of layout components are: \colorbox{question}{Question}, \colorbox{answer}{Answer}, \colorbox{header}{Header}, \colorbox{other}{Other}. Our Doc-GCN produced the best recognition results, especially for the \textit{Other} components.}
        \label{fig:funsd_addition_case}
\end{figure*}

\begin{figure*}[ht!]
    \hspace*{-0.5cm}
    \captionsetup[subfigure]{aboveskip=-10pt,belowskip=-4pt,justification=centering}
     \centering
     \begin{subfigure}[b]{0.24\textwidth}
         \centering
         \includegraphics[width=4cm]{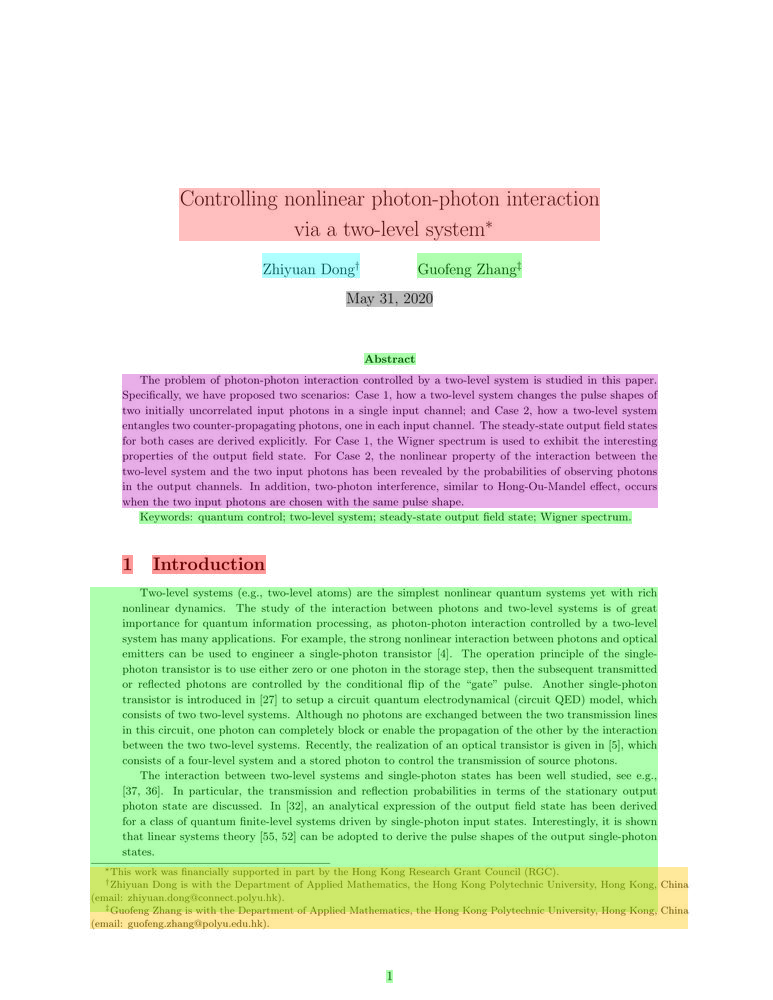}
         \caption{\textbf{Ground Truth}}
         \label{fig:case_study_GT2}
     \end{subfigure}
     \hspace*{-0.2em}
     \begin{subfigure}[b]{0.24\textwidth}
         \centering
         \includegraphics[width=4cm]{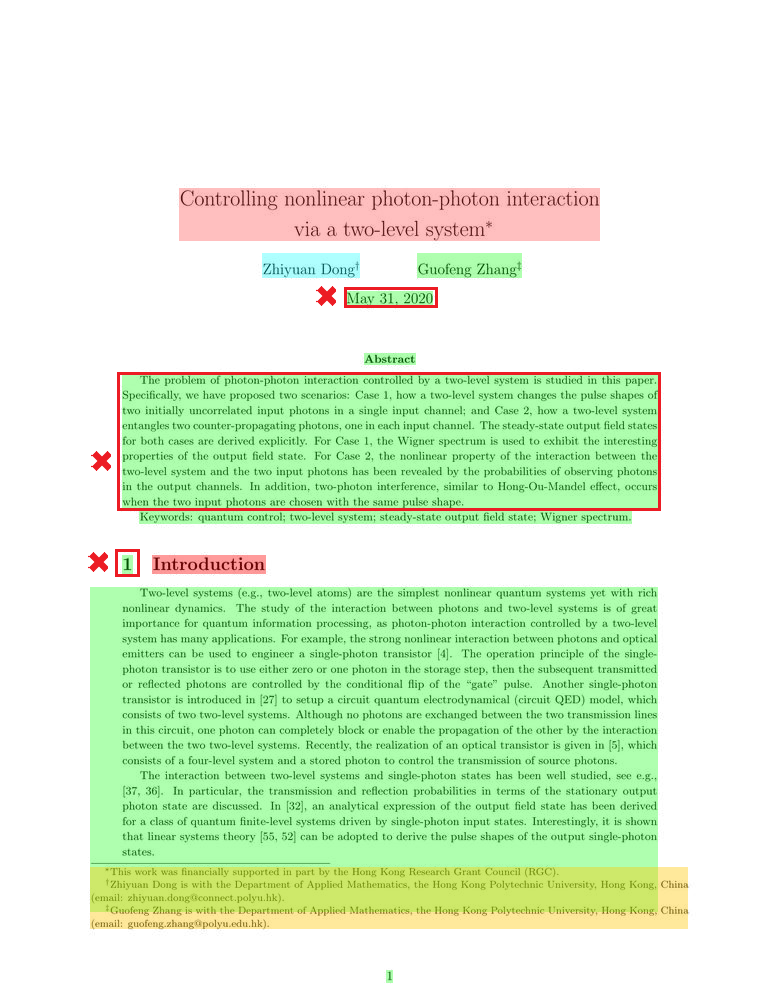}
         \caption{\textbf{BERT-base}}
         \label{fig:case_study_bert3}
     \end{subfigure}
     \hspace*{-0.2em}
     \begin{subfigure}[b]{0.24\textwidth}
         \centering
         \includegraphics[width=4cm]{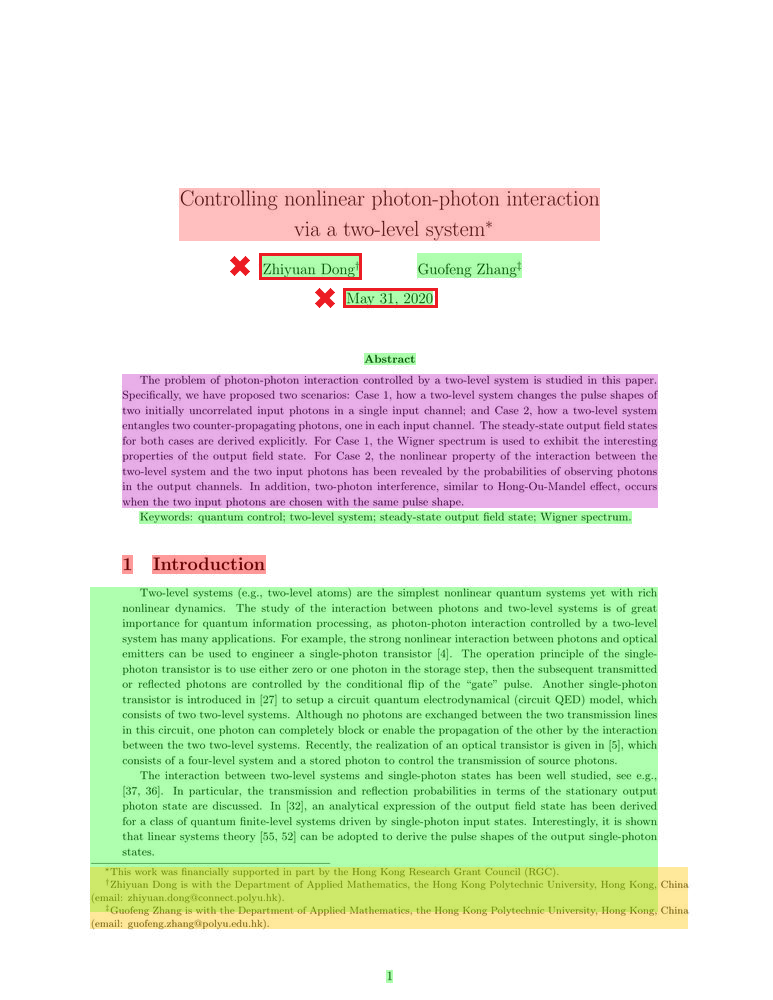}
         \caption{\textbf{RoBERTa-base}}
         \label{fig:case_study_layoutlm2}
     \end{subfigure}
     \hspace*{-0.2em}
     \begin{subfigure}[b]{0.24\textwidth}
         \centering
         \includegraphics[width= 4cm]{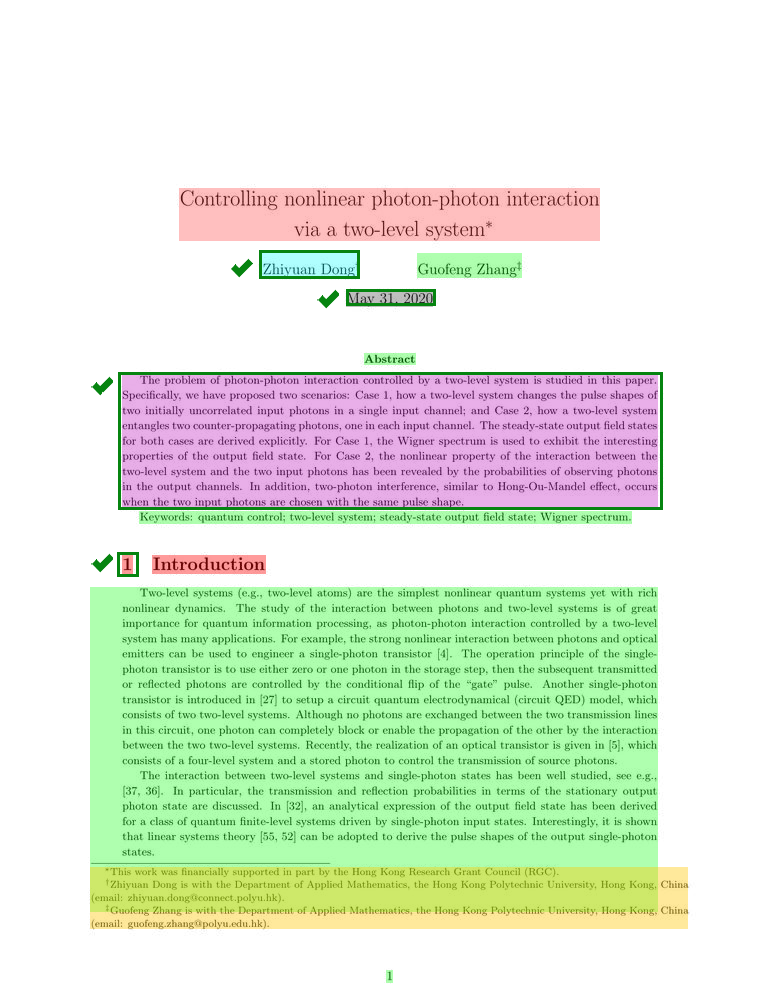}
         \caption{\textbf{Doc-GCN}}
         \label{fig:docbank_ourmodel2}
     \end{subfigure}
        \caption{Visualization of Top3 models (BERT-base, RoBERTa-base and Doc-GCN) for a DocBank PDF page. The color of layout component labels are: \colorbox{abstract}{Abstract}, \colorbox{author}{Author}, \colorbox{caption}{Caption}, \colorbox{date}{Date}, \colorbox{equation}{Equation}, \colorbox{figure_doc}{Figure}, \colorbox{footer}{Footer}, \colorbox{list_doc}{List}, \colorbox{paragraph}{Paragraph}, \colorbox{reference}{Reference}, \colorbox{section}{Section}, \colorbox{table_doc}{Table}, \colorbox{title_doc}{Title}. Our Doc-GCN classified all layout components accurately.}
        \label{fig:vis_docbank}
\end{figure*}

\begin{figure*}[ht!]
    \hspace*{-0.5cm}
    \captionsetup[subfigure]{aboveskip=-10pt,belowskip=-4pt,justification=centering}
     \centering
     \begin{subfigure}[b]{0.25\textwidth}
         \centering
         \includegraphics[width=4.1cm]{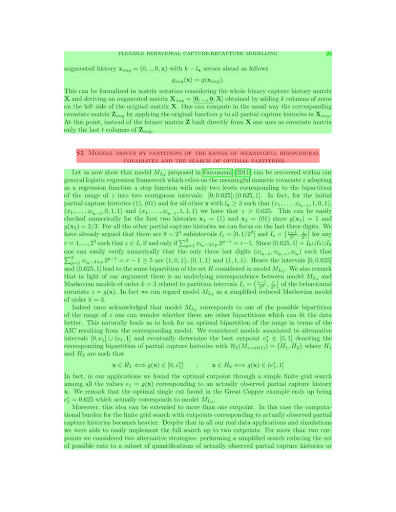}
         \caption{\textbf{Ground Truth}}
         \label{fig:GT_DocBANK_appendix}
     \end{subfigure}
     \hspace*{-0.2em}
     \begin{subfigure}[b]{0.25\textwidth}
         \centering
         \includegraphics[width=4.1cm]{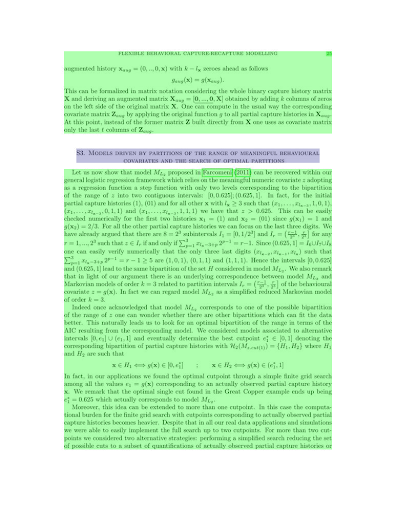}
         \caption{\textbf{BERT-base}}
         \label{fig:BERT_DocBANK_appendix}
     \end{subfigure}
     \hspace*{-0.2em}
     \begin{subfigure}[b]{0.25\textwidth}
         \centering
         \includegraphics[width=4.1cm]{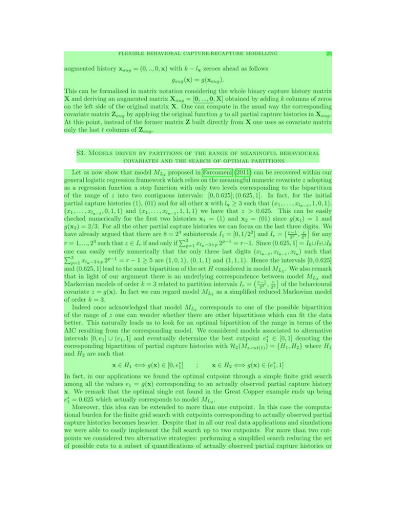}
         \caption{\textbf{RoBERTa-base}}
         \label{fig:Roberta_DocBANK_appendix}
     \end{subfigure}
     \hspace*{-0.2em}
     \begin{subfigure}[b]{0.25\textwidth}
         \centering
         \includegraphics[width= 4.1cm]{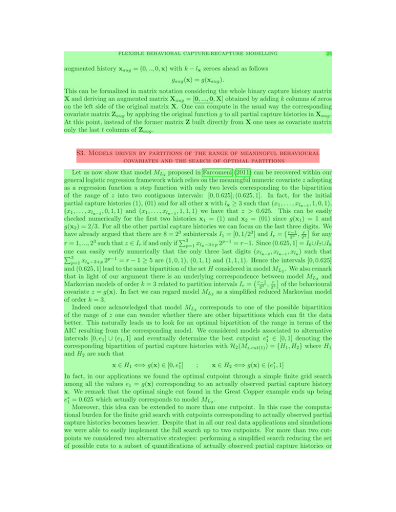}
         \caption{\textbf{Doc-GCN}}
         \label{fig:DocGCN_DocBANK_appendix}
     \end{subfigure}
        \caption{Visualization of Top3 models for a \textbf{DocBank} PDF page. The color of layout component labels are: \colorbox{abstract}{Abstract}, \colorbox{author}{Author}, \colorbox{caption}{Caption}, \colorbox{date}{Date}, \colorbox{equation}{Equation}, \colorbox{figure_doc}{Figure}, \colorbox{footer}{Footer}, \colorbox{list_doc}{List}, \colorbox{paragraph}{Paragraph}, \colorbox{reference}{Reference}, \colorbox{section}{Section}, \colorbox{table_doc}{Table}, \colorbox{title_doc}{Title}. Our Doc-GCN classified all layout components accurately.}
        \label{fig:docbank_addition_case}
\end{figure*}

We provided one additional quality result on the document of PubLayNet (Figure~\ref{fig:publaynet_addition_case}), FUNSD (Figure~\ref{fig:vis_funsd},\ref{fig:funsd_addition_case}) and DocBank (Figure~\ref{fig:vis_docbank},\ref{fig:docbank_addition_case}),respectively. As shown in Figure~\ref{fig:publaynet_addition_case}, both RoBERTa-base and Faster-RCNN incorrectly recognize the \textit{List} into \textit{Text}, whereas our Doc-GCN successfully recognize all segments for Publaynet Dataset. 

As shown in Figure~\ref{fig:vis_funsd}, Doc-GCN also shows its remarkable performance by correctly recognising the top-right component highlighted by the green tick in Figure~\ref{fig:funsd_ourmodel} while the other two baseline models wrongly classified this component into \textit{Answer}. Additionally, ours detected \textit{Other} components such as page number, company name, and date time correctly. This is another highlight of the great potential of our multi-modal aspects, even with the small-sized of the training dataset. Similar trend also can be found in Figure~\ref{fig:funsd_addition_case}.

Figure~\ref{fig:vis_docbank}, a dataset with the scientific academic papers, shows that Doc-GCN produced all components correctly. However, BERT could not detect \textit{Date} (published date) and \textit{Abstract}, which are the most important components of the publication. RoBERTa also incorrectly recognised the \textit{Date} (published date) and \textit{Author}. Thus it is clear to see that our Doc-GCN consistently works well in detecting paragraphs and equations. As well as, in Figure~\ref{fig:docbank_addition_case}, only our Doc-GCN can get the exactly same results as the ground truth.

\end{document}